%% file: main.tex
\documentclass{Interspeech}
\usepackage{multirow}
\usepackage{amssymb}
\usepackage{pifont}
\usepackage{mathtools}
\usepackage[dvipsnames]{xcolor}
\usepackage{textcomp}
\usepackage{tikz}
\usepackage{pgfplotstable}
\usetikzlibrary{patterns.meta}
\usepackage{graphicx}
\usepackage{standalone}
\usepackage{subcaption}
\usepackage{pgfplots}
\usepackage{adjustbox}
\usepackage{pgfplotstable}

\graphicspath{{figures/}}

\pgfplotsset{
    my ybar legend/.style={
        legend image code/.code={
            \draw [##1] (0cm,-0.6ex) rectangle +(2.5em,1.5ex);
        },
        row sep=0.5cm
    },
}

\def\b{{\phantom{$^{\text{\tiny{+}}}$}}}

\def\bbb{{\phantom{$^{\text{\tiny{+++}}}$}}}

\interspeechcameraready

\renewrobustcmd{\bfseries}{\fontseries{b}\selectfont}
\renewrobustcmd{\boldmath}{}
\newrobustcmd{\B}{\bfseries}

\newsavebox\CBox

\makeatletter

\renewcommand{\section}{\@startsection
   {section}%
   {1}%
   {}%
   {0.15\baselineskip}%
   {0.2\baselineskip}%
   {}}%

\renewcommand{\subsection}{\@startsection
  {subsection}%
  {2}%
  {}%
  {0.2\baselineskip}%
  {0.2\baselineskip}%
  {}}%

\renewcommand{\subsubsection}{\@startsection
  {subsubsection}%
  {3}%
  {}%
  {-0.2\baselineskip}%
  {0.1\baselineskip}%
  {}}%

\g@addto@macro\normalsize{%
  \setlength\abovedisplayskip{3pt plus 2pt minus 1pt}
  \setlength\belowdisplayskip{3pt plus 2pt minus 1pt}
  \setlength\abovedisplayshortskip{2pt plus 2pt minus 1pt}
  \setlength\belowdisplayshortskip{2pt plus 2pt minus 1pt}
}

\setlist{
    itemsep=0pt,
    parsep=1pt plus 1pt minus 1pt,
    topsep=1pt plus 1pt minus 1pt,
    partopsep=0pt
}

\title{Dynamic Acoustic Model Architecture Optimization in Training for ASR}

\author[affiliation={1,2}]{Jingjing}{Xu}
\author[affiliation={1}]{Zijian}{Yang}
\author[affiliation={1,2}]{Albert}{Zeyer}
\author[affiliation={2}]{Eugen}{Beck}
\author[affiliation={1,2}]{Ralf}{Schlüter}
\author[affiliation={1,2}]{Hermann}{Ney}

\affiliation{Machine Learning and Human Language Technology Group}{RWTH Aachen University}{Germany}
\affiliation{}{AppTek GmbH}{Germany}
\email{\{jxu, zeyer, zyang, schlueter\}@ml.rwth-aachen.de, ebeck@apptek.com}
\keywords{speech recognition, dynamic architecture optimization, CTC}

\usepackage{comment}

\begin{document}

\maketitle

\begin{abstract}
Architecture design is inherently complex.
Existing approaches rely on either handcrafted rules, which demand
extensive empirical expertise, or automated methods like neural architecture
search, which are computationally intensive.
In this paper, we introduce DMAO, an architecture optimization
framework that employs a grow-and-drop strategy to automatically
reallocate parameters during training.
This reallocation shifts resources from less-utilized areas to those parts of
the model where they are most beneficial.
Notably, DMAO only introduces negligible training overhead at a given model
complexity.
We evaluate DMAO through experiments with CTC on \textit{LibriSpeech},
\textit{TED-LIUM-v2} and \textit{Switchboard} datasets.
The results show that, using the same amount of training resources, our proposed
DMAO consistently improves WER by up to $\sim6\%$ relatively across various
architectures, model sizes, and datasets.
Furthermore, we analyze the pattern of parameter redistribution and uncover
insightful findings.
\end{abstract}

\section{Introduction \& Related Work}
In the last decade, self-attention-based architectures such as Transformer \cite{vaswani2017transformer},
Conformer \cite{burchi2021conformer}, and E-branchformer \cite{kim2022ebranchformer}
have revolutionized the automatic speech recognition (ASR) field,
significantly boosting the performance of acoustic models.
For convenience and efficient scalability, identical neural blocks are stacked
together to construct the acoustic encoder.
However, this may result in inefficient use of parameters, as research
\cite{shim2022role, pasad2021analysis, shim2022attn} has shown that the model
may perform different roles at varying depths, and such behavior is data-dependent.
The basic components of the model architecture, such as feed-forward,
convolutional, and attention layers, are specialized for different tasks.
The question of how to design a better data-specific model architecture by
optimally distributing the parameters of different types at various depths based
 on their usefulness remains under-explored.

The recently proposed Zipformer \cite{yao2024zipformer} divides the layers into
stacks, with each stack having a different embedding dimension.
However, the authors do not provide an explanation for the choice of these specific dimensions.
In the works \cite{mehta2021delight,mehta2024openelm}, a block-wise scaling
approach based on hand-crafted rules is employed to adjust the depth and width
of blocks.
However, the hyper-parameters governing the scaling process still require tuning
to achieve optimal performance, thereby introducing additional training efforts.
Neural architecture search (NAS) can be used to discover efficient architectures
that outperform hand-designed ones and has been applied in
\cite{liu2022nas, mehrota2021nas, liu2021nas} for the ASR task.
However, NAS approaches often demand significantly more time and computational
resources due to the extensive exploration of the architecture space and repeated
evaluations of candidate models.
These previous solutions are non-trivial.
They either demand extensive expertise from researchers or require additional
training efforts.
Therefore, in this work, we aim to find a solution that dynamically optimizes
the model architecture within a fixed resource budget during training,
without incurring additional overhead.

The grow-and-drop paradigm has been utilized for training sparse networks
from scratch \cite{mocanu2018scalable, evci2020sparsenet, evci2022sparsenet,mostafa2019sparse}.
In those works, training begins with a randomly initialized sparse network with
predefined sparsity, but the optimal architecture remains unknown.
During the grow phase, neurons or connections are either added randomly or based
on certain strategies, allowing the model to explore a variety of architectural
configurations.
After the growth phase, the network deactivates parameters that are considered
unimportant.

In this work, we extend the grow-and-drop paradigm to dense network training.
We propose a simple Dynamic Model Architecture Optimization (DMAO)
framework that utilizes the grow-and-drop strategy to efficiently redistribute
parameters during the training process.
To enable flexible adaptation, we partition the dense model into finer groups of
 parameters.
We define and compare various metrics to assess the importance of these parameter
groups based on their contribution to the ASR model's performance.
The parameter groups are ranked according to these metrics, after which we grow
the top-ranked groups and remove the lowest-ranked ones.
In this way, we enhance the model’s capacity by reallocating resources from
underutilized areas to those where they are most needed.
We evaluate our approach by conducting experiments with the connectionist temporal
classification (CTC) \cite{graves2006ctc} model on \textit{LibriSpeech},
\textit{TED-LIUM-v2} and \textit{Switchboard} datasets.
The results demonstrate that DMAO consistently improves word-error-rate (WER) by up to $\sim$6\%
relative across different architectures (e.g. Conformer, E-Branchformer),
model sizes, and datasets.
We also investigate the optimal schedule for applying DMAO during training.
Furthermore, we analyze how the parameter distribution changes with DMAO and provide
possible explanations.

\section{Dynamic Model Architecture Optimization}
In this section, we present our approach for dynamically optimizing the acoustic
model architecture during training.
The core idea is to redistribute the parameters from parts of the model where
they are least useful to those where they are most critical, thereby enhancing
the model's capacity
while maintaining a fixed budget for specific resource constraints.
To enable more flexible global adjustments, we begin with partitioning the model
into smaller parameter groups.
We then rank these groups based on their importance, as computed using the
metrics described in Sec.~\ref{sec:compute_importance_score}.
The parameters are reallocated using the grow-and-drop paradigm, i.e., by
removing the least important ones and duplicating the most important ones.
The remaining parameters are left unchanged, and training proceeds with the
updated model.

\subsection{Model Partition}
\label{sec:model_partition}
Conformer \cite{gulati2020conformer} and E-branchformer
\cite{kim2022ebranchformer}, which are among the most prevalent encoder
architectures for ASR tasks, are used in this work.
Conformer consists of feed-forward network (FFN), multi-head self-attention module
(MHSA) and convolutional module (Conv).
FFN has two weight matrices $W_{\textrm{ff}_1},W_{\textrm{ff}_2}^T \in \mathbb{R}^{d_{\textrm{model}} \times d_{\textrm{ff}}}$,
$d_{\textrm{model}}$ is the model dimension and $d_{\textrm{ff}}$ denotes the inner dimension of FFN.
We partition the FFN into $C$ groups,with sub-matrices
 $W_{\textrm{ff}_1}^{c}, {W_{\textrm{ff}_2}^c}^T \in \mathbb{R}^{d_{\textrm{model}} \times \frac{d_{\textrm{ff}}}{C}}$.
MHSA has $H$ heads and consists of four projection matrices, $W_k, W_q, W_v, W_o^T \in
\mathbb{R}^{d_{\textrm{model}}\times{(d_h \cdot H)}}$, $d_h$ is the dimension per head.
We partition the MHSA into $H$ groups, each with sub-matrices
$W_k^{h}, W_q^{h}, W_v^{h}, {W_o^h}^T \in \mathbb{R}^{d_{\textrm{model}}\times d_h}$.
Conv consists of weights $W_{\textrm{in}} \in \mathbb{R} ^{d_{\textrm{model}} \times 4d_{\textrm{model}}}$,
$W_{\textrm{conv}} \in \mathbb{R} ^{k \times k \times 2d_{\textrm{model}} \times 1}$, and
 $W_{\textrm{out}} \in \mathbb{R} ^{2d_{\textrm{model}} \times d_{\textrm{model}}}$.
Likewise, we partition one Conv into $M$ groups, with sub-matrices
$W_{\textrm{in}}^m \in \mathbb{R} ^{d_{\textrm{model}} \times \frac{4d_{\textrm{model}}}{M}}$,
$W_{\textrm{conv}}^m \in \mathbb{R} ^{k \times k \times \frac{2d_{\textrm{model}}}{M} \times 1}$
and  $W_{\textrm{out}}^m \in \mathbb{R} ^{\frac{2d_{\textrm{model}}}{M} \times d_{\textrm{model}}}$.
This partition allows for easy adjustment of the hidden dimensions in each module.
The E-branchformer can be partitioned into a similar manner.
Suppose the encoder has $L$ layers, we partition the model into
$(2C+H+M) \times L$ groups.

\subsection{Ranking Based on Importance Scores}
\label{sec:compute_importance_score}
We define several metrics to assess the importance of parameter
groups.
These scores are used to rank parameter groups according to their
contribution to ASR performance.
Additionally, they should be efficient to calculate and easily accessible,
ensuring minimal computational overhead.
Suppose the parameter group has $N$ weight elements $w_i \in \mathbb{R}, i=\{1, ..., N\}$. \\
\textbf{Magnitude} refers to the absolute value of a weight and are widely used
 as metric for pruning ASR models \cite{cheng2021zeroout, jim2021magnitude}.
  Weights with larger values after applying a norm to their magnitudes are
  considered more important, as they have a greater influence on the activation and final output
   \cite{han2015magnitude, frankle2019lth}.
  To calculate the importance score at training step t, we average the
  magnitudes of all weights in the parameter group.
  Additionally, to make the score more stable, we use exponential smoothing,
  as in \cite{yang2023firsttaylor}, where $\alpha$ is a constant smoothing factor.
  \scalebox{1}{\parbox{1\linewidth}{%
  \begin{align}
    s_t = (1 - \alpha) s_{t-1} + \alpha \left( \frac{1}{N} \sum_{i=1}^{N} |w_i| \right)
    \label{eq:magnitude}
  \end{align}
  }}
  \textbf{Gradient} represent the rate of change of the loss function with respect
  to the model's weights \cite{ding2019gradient,bekal2021metric}.
  Weights with small absolute gradients have a smaller impact on reducing the
  model's loss and are therefore less influential in the training process.
  To assess the importance of each parameter group, we compute the averaged $L_2$ norm
   of the gradients across all weights in that group and apply exponential smoothing as
   in Eq.~(\ref{eq:magnitude}) as well.
  \scalebox{1}{\parbox{1\linewidth}{%
  \begin{align}
  s_t = (1 - \alpha) s_{t-1} + \alpha \frac{1}{N} \sqrt{\sum_{i=1}^N \left( \frac{\partial \mathcal{L}_{\textrm{ASR}}}{\partial w_i} \right)^2}
  \label{eq:gradient}
  \end{align}
  }}
  \textbf{First-order Taylor Approximation} provides a linear approximation of
  a function around a given point.
  We compute the importance score of an individual weight $w_i$ using the
  first-order Taylor approximation to estimate the loss difference when $w_i$ is
  removed or equivalently set to zero, as in \cite{michel2019firsttaylor, yang2023firsttaylor}.
  Compared to using gradients alone, this approach also takes the weight value into
  account, ensuring that both the size and effect of the weights are accounted
  for the decision-making process.
  For parameter groups, we compute the importance score using the same averaged
   $L_2$-norm as in Eq.~(\ref{eq:gradient}).
   \scalebox{1}{\parbox{1\linewidth}{%
   \begin{align}
   s_t = (1 - \alpha) s_{t-1} + \alpha \frac{1}{N} \sqrt{\sum_{i=1}^N \left( \frac{\partial \mathcal{L}_{\textrm{ASR}}}{\partial w_i} w_i \right)^2}
   \label{eq:first_taylor}
   \end{align}
   }}
  \textbf{Learnable Score}
  Inspired by movement pruning \cite{sanh2019movementpruning, lagunas2021blockmovement},
  we assign a learnable score $s$ to each parameter group to indicate its importance.
  We directly scale all weight elements in that group as
  \( w_i' = w_i \times s \).
  After each adaptation, the scales of all parameter groups (i.e., $s$) are reset to 1 to
  ensure a fair comparison in the next round.
  Empirically, we observe that the model may diverge if the scaling effect is
  removed.
Therefore, we randomly sample training steps with a probability of 0.5, i.e.,
for half of the training steps, unscaled weights are used, and for the other half,
scaled weights are applied.
  In this way, the model remains robust to the scaling effect.

\subsection{Architecture Optimization}
\begin{figure}
  \centering
  \includegraphics[scale=0.33]{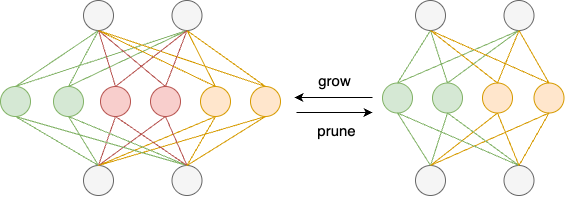}
  \caption{Illustration of grow/prune parameter groups in an FFN module. Grey nodes indicate input/output. The parameter groups (green, red, yellow) include hidden nodes and their connections. Selecting the green group for doubling adds $\frac{d_{\textrm{ff}}}{C}$ new hidden nodes and their connections (in red),
  updating FFN weight matrices to $W_{\textrm{ff}_1},W_{\textrm{ff}_2}^T \in \mathbb{R}^{d_{\textrm{model}} \times \frac{C+1}{C}d_{ff}}$.
 Pruning the red group removes its nodes and corresponding connections.}
  \label{fig:grow_and_prune}
  \vspace{-1em}
\end{figure}

The adaptation is performed iteratively over \( I \) iterations during training.
Let \( T_{\text{end}} \) denote the training step at which adaptation stops,
and \( \Delta T \) the number of training steps per iteration.
At the end of each \( i^{\text{th}} \) iteration, i.e., at the \( i\Delta T \)
training step, we use the updated importance scores to rank the parameter groups
and optimize the model architecture.
Let $\delta \in [0, 0.5]$ denote the adaptation ratio, then we optimize the
architecture by removing the bottom $\frac{\delta}{I}$ fraction of the parameters and
doubling the top $\frac{\delta}{I}$ fraction.
As a result, a total of $2\delta$ parameters are modified.
The grow-and-prune paradigm of FFN is illustrated in Figure~\ref{fig:grow_and_prune} and
can be easily analogized to all Conformer modules.
With the partitioning design described in Sec.~\ref{sec:model_partition},
only the hidden dimension of each module is modified, while the input and output
dimensions remain unchanged.
This design enables each parameter group to be independently doubled or pruned
without altering the shapes of other groups.
The initialization of the newly introduced weights is investigated in
Sec.~\ref{sec:weight_init}.
Alternatively, we can use different resource constraints,
such as the number of FLOPs, as used in \cite{xu2025efficient}.

\section{Experiments}
\subsection{Experimental Setup}
We conduct the experiments on the 960h \textit{LibriSpeech} (\textit{LBS})
corpus \cite{vassil2015lbs960}, the 200h \textit{TED-LIUM-v2} (\textit{TED-v2})
corpus \cite{rousseau2014tedlium} and the 300h \textit{Switchboard} (\textit{SWB})
corpus \cite{Godfrey1992SWB}.
We use a phoneme-based CTC model following the setup in \cite{xu2023dynamic} as
baseline,  with log Mel-filterbank features as the input.
It has a VGG front-end and 12 Conformer/E-branchformer blocks.
The output labels are 79 end-of-word augmented phonemes.
SpecAugment \cite{zhou2020specaug} is applied for data augmentation.
For Conformer, the number of attention heads $H$ is $\frac{d_{\textrm{model}}}{64}$ and
$d_{\textrm{ff}}=4 \times d_{\textrm{model}}$.
For E-branchformer, the inner dimension of local extractor $d_{\textrm{inter}}$ is set to
$6 \times d_{\textrm{model}}$.
In terms of model partition, we set both $C$ and $M$ to 4.
We train all models for the same number of epochs: 50 epochs for \textit{SWB}
and \textit{TED-v2}, and 30 epochs for \textit{LBS}.
The one-cycle learning rate scheduler \cite{zhou2022oclr} is used, with the
learning rate increasing from \(4 \times 10^{-6}\) to \(4 \times 10^{-4}\) for
45\% of the training, decreasing for the next 45\%, and decaying to \(10^{-7}\)
in the final 10\%.
We use RETURNN \cite{zeyer2018returnn} to train the acoustic models and RASR
\cite{wiesler2014rasr} for recognition.
In inference, we apply Viterbi decoding with a 4-gram word-level language model.
To further reduce computation, the smoothed scores in Eq.~(1-3) are computed and
updated every 1000 steps.
All our config files and code to reproduce the results can be found online\footnote{\scriptsize https://github.com/rwth-i6/returnn-experiments/tree/master/2025-dynamic-model-architecture-optimization}.

\subsection{Experimental Results}

\subsubsection{Overall Results with Conformer Encoder}
\begin{figure}
  \centering
  \includegraphics[scale=0.39]{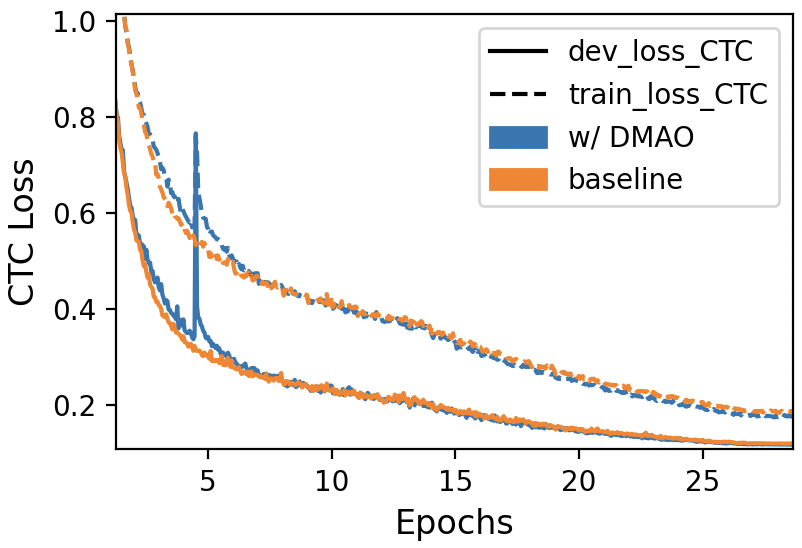}
  \caption{Comparison of training loss between the baseline and the model w/
  DMAO on \textit{LibriSpeech} dataset.}
  \label{fig:loss_plot}
  \vspace{-2em}
\end{figure}

We present the overall results with Conformer across different model sizes in
Table~\ref{tab:overall_results_conformer}.
The optimal DMAO setting including the selection of importance metrics
is used, ablation study results are presented in the following sections.
The results show that DMAO improves WER performance across all model sizes and datasets,
 achieving up to a ~6\% relative improvement.
Furthermore, within each size category, the `retrain' model with the
optimized architecture outperforms the baseline model.
To ensure a fair comparison, all training factors, except for the model
architecture, remain consistent.
This suggests that the updated architecture provides greater model capacity.
Moreover, when compared to retraining with optimized architecture, the model
with DMAO yields comparable or even slightly better results, indicating that
directly applying adaptation during training does not disrupt model convergence.
Figure \ref{fig:loss_plot} plots the training loss for both the baseline model
and the model with DMAO.
After the architecture update, we observe a sharp increase in training loss,
as the model's weights and parameter distribution are altered. However, the loss
decreases quickly, indicating that the model recovers rapidly.
Following this, the training loss of the DMAO model remains slightly lower than
the baseline, verifying our assumption that the updated model has greater capacity.
To assess the reliability of the improvement, we compute
bootstrap-estimated probabilities of WER reduction on each test set
using the implementation from \cite{bisnai2004bootstrp}.
Absence of notation $^{\text{\tiny{+}},\text{\tiny{++}},\text{\tiny{+++}}}$
signifies a confidence level below 90\%.
The results indicate strong confidence that DMAO improves WER compared to the baseline on both the \textit{LBS} and \textit{SWB} datasets.

\begin{table}
  \centering
  \caption{Overall WERs [\%] results using Conformer as the encoder architecture
  across different model sizes on three datasets. The baseline model does not use DMAO.
  The `w/ DMAO' model applies model architecture optimization once at 20\%
  (15\% for \textit{LBS}) of the total training steps, with the adaptation ratio
  $\delta$ set to 0.15, using the first-order Taylor approximation as the metric.
  The `retrain' model uses the final architecture from the `w/ DMAO'
  model but is trained from scratch.}
  \label{tab:overall_results_conformer}
  \setlength{\tabcolsep}{0.3em}
  \scalebox{0.9}{\begin{tabular}{|c|l||c|c|c|c|c|c|}
    \hline
    \multirow{2}{*}{\shortstack{Params.\\ $[\text{M}]$}}& \multirow{2}{*}{Model} & \multicolumn{2}{c|}{TED-v2} & \multicolumn{2}{c|}{SWB} & \multicolumn{2}{c|}{LBS} \\ \cline{3-8}
    & & dev & test & \shortstack{Hub\\5'00} & \shortstack{Hub\\5'01} & \shortstack{dev-\\other} & \shortstack{test-\\other} \\ \hline \hline
    \multirow{3}{*}{19.0} & baseline & 7.8 & 8.2\b & 14.8 & 13.4\bbb & 7.6 & 7.9\bbb \\ \cline{2-8}
    & w/ DMAO & \textbf{7.6} & \textbf{7.7}\b & \textbf{14.2} & \textbf{12.8}$^{\text{\tiny{+++}}}$ &
    \textbf{7.3}& \textbf{7.8}$^{\text{\tiny{+++}}}$ \\ \cline{2-8}
    & \hspace{3mm}+retrain & 7.7 & 7.9$^{\text{\tiny{+}}}$ & 14.2 & 13.1$^{\text{\tiny{+++}}}$ & 7.5 & 7.9$^{\text{\tiny{+++}}}$\\ \hline \hline
    \multirow{3}{*}{42.1} & $\text{baseline}^{\dag}$ & 7.7 & \textbf{7.9}\b & 14.0 & 12.8\bbb & 7.0 & 7.5\bbb \\ \cline{2-8}
    & w/ DMAO$^\ddag$ & \textbf{7.3} & \textbf{7.9}\b & \textbf{13.5} & \textbf{12.3}$^{\text{\tiny{+++}}}$ & 6.8 & \textbf{7.2}$^{\text{\tiny{+++}}}$ \\ \cline{2-8}
    & \hspace{3mm}+retrain & 7.4 & \textbf{7.9}\b & 13.9 & 12.5$^{\text{\tiny{++}}}$\b & \textbf{6.7} & \textbf{7.2}$^{\text{\tiny{+++}}}$\\ \hline \hline

    \multirow{3}{*}{72.8} & baseline & 7.6 & 7.7\b & 13.9 & 12.6\bbb & 6.8 & 7.1\bbb  \\ \cline{2-8}
    & w/ DMAO & 7.4 & 7.7\b & 13.8 & 12.2$^{\text{\tiny{++}}}$\b & \textbf{6.6} & \textbf{6.9}$^{\text{\tiny{++}}}$\b \\ \cline{2-8}
    & \hspace{3mm}+retrain & \textbf{7.3} & \textbf{7.5}\b & \textbf{13.2} & \textbf{11.8}$^{\text{\tiny{+++}}}$ &
    \textbf{6.6} & 7.0\bbb\\ \hline
  \end{tabular}}
  \leftline{{\scriptsize $^\dag$ used as the baseline for Table~\ref{tab:scoring_metrics_comparison}, Table~\ref{tab:initialization_ablation}
  and Table~\ref{tab:adaptation_schedule_ablation}. $^\ddag$ used to plot Figure \ref{fig:ratio_after_update}.}}
  \leftline{{\scriptsize $^{\text{\tiny{+}},\text{\tiny{++}},\text{\tiny{+++}}}$ denotes probability of
  improvement $>90\%$, $>95\%$, and $>99\%$.}}
\end{table}

\subsubsection{Importance Score Metrics Comparison}
In Table~\ref{tab:scoring_metrics_comparison}, we compare the ASR results using
different scoring metrics for ranking parameter groups in dynamic model
architecture adaptation.
Except for the magnitude-based scoring metric, all other metrics lead to WER
improvements over the baseline, highlighting the importance of choosing a good
metric.
Among the metrics, the first-order Taylor approximation achieves the best
performance, showing a $\sim4\%$ improvement over the baseline on both the
\textit{SWB} Hub5'01 and \textit{LBS} test-other sets.
Additionally, the results suggest that exponential smoothing improves WER,
making the scores more reliable.

\begin{table}
  \centering
  \setlength{\tabcolsep}{0.3em}
  \caption{Comparison of WERs [\%] using different metrics and various values
  of the smoothing factor $\alpha$ to compute importance scores for parameter
  groups.
  The baseline model does not use dynamic architecture adaptation, whereas the
  other models apply it once at 20\% (15\% for LBS) of the total training steps with adaptation
  ratio $\delta$ 0.15.
  The baseline is $^\dag$ from Table~\ref{tab:overall_results_conformer}.
 }
 \label{tab:scoring_metrics_comparison}
  \scalebox{0.9}{\begin{tabular}{|c|c|c|c|c|c|}
    \hline
    \multirow{2}{*}{Metric} & \multirow{2}{*}{\shortstack{Smooth \\ factor $\alpha$}} & \multicolumn{2}{c|}{SWB} & \multicolumn{2}{c|}{LBS} \\ \cline{3-6}
    & & Hub5'00 & Hub5'01 & dev-o & test-o \\ \hline
    baseline & - & 14.0 & 12.8 & 7.0 & 7.5 \\ \hline
    magnitude & \multirow{3}{*}{0.9} & 14.0 & 12.8 & 7.7 & 7.9 \\ \cline{1-1} \cline{3-6}
    gradient & & 13.6 & 12.3 & 7.0 & \textbf{7.2} \\ \cline{1-1} \cline{3-6}
    \multirow{4}{*}{\shortstack{first order \\ Taylor}} & & \textbf{13.5}
    & 12.3 & \textbf{6.8} & \textbf{7.2}\\ \cline{2-6}
    & 1 & 13.7 & 12.3 & 7.1 & 7.5 \\ \cline{2-6}
    & 0.7 & 13.8 & 12.5 & 6.9 & 7.3  \\ \cline{2-6}
    & 0.5 & 13.6 & 12.3 & 7.0 & 7.3 \\ \hline
    learnable score & - & 13.6 & \textbf{12.2} & \textbf{6.9} & 7.3\\ \hline
  \end{tabular}}
\end{table}

\subsubsection{Initialization Comparison for Newly Introduced Weights}
\label{sec:weight_init}
In Table~\ref{tab:initialization_ablation}, we investigate different
initialization strategies for the newly introduced weights after each model architecture update.
The results show that directly copying the weights from the top $\delta$
parameters yields the best performance.

\begin{table}
  \centering
  \setlength{\tabcolsep}{0.2em}
  \caption{WERs [\%] comparison of different initialization for the newly
  introduced weights after each architecture optimization.
  The baseline is $^\dag$ and `copy' is $^\ddag$ from Table~\ref{tab:overall_results_conformer}.
  `copy' refers to using the exact same weights from the top $\delta$ parameters,
  while `copy + noise' adds extra Gaussian noise with a mean of 0 and
  a standard deviation of 0.01.}
  \label{tab:initialization_ablation}
  \scalebox{0.9}{\begin{tabular}{|l|c|c|c|c|}
    \hline
    \multirow{2}{*}{Initialization} & \multicolumn{2}{c|}{SWB} & \multicolumn{2}{c|}{LBS} \\ \cline{2-5}
    & Hub5'00 & Hub5'01 & dev-other & test-other \\ \hline
    baseline w/o DMAO & 14.0 & 12.8 & 7.0 & 7.5 \\ \hline
    copy & \textbf{13.5} & \textbf{12.3} & \textbf{6.8} & \textbf{7.2}\\ \hline
    copy + noise & 13.7 & \textbf{12.3} & 6.9 & \textbf{7.2} \\ \hline
    random & 13.6 & \textbf{12.3} & 7.0 & 7.4 \\ \hline
  \end{tabular}}
\end{table}

\subsubsection{Ablation Study of DMAO schedule}
Table~\ref{tab:adaptation_schedule_ablation} presents the results of an ablation
study on the DMAO schedule.
We observe that, applying DMAO too late in the training process can degrade
performance, possibly because the model struggles to recover from the updates.
On the other hand, applying it too early is suboptimal, as the scores computed
in the initial training steps may be suboptimal.
A good balance is observed when applying DMAO at around 20\% of the total
training steps, during a phase where the model has not yet plateaued but the loss
is no longer changing rapidly.
Applying DMAO once tends to outperform multiple iterations, as multiple architecture
updates may introduce excessive disruption.
Lastly, an adaptation ratio \(\delta\) between 0.15 and 0.25 appears to be effective.

\begin{table}
  \centering
  \setlength{\tabcolsep}{0.45em}
  \caption{WERs [\%] of applying DMAO at different training
  stages, with varying adaptation ratios and number of iterations (first-order
  Taylor approximation as metric).
  $T_{\textrm{total}}$ denotes the total number of training steps.
  The baseline is $^\dag$ from Table~\ref{tab:overall_results_conformer}.
  }
  \label{tab:adaptation_schedule_ablation}
  \scalebox{0.9}{\begin{tabular}{|c|c|c|c|c|c|c|}
    \hline
    \multirow{2}{*}{$\frac{T_{\textrm{end}}}{T_{\textrm{total}}}$} &  \multirow{2}{*}{\shortstack{Update \\ ratio $\delta$}} & \multirow{2}{*}{\shortstack{Num. \\ iters. I}} & \multicolumn{2}{c|}{TED-v2} & \multicolumn{2}{c|}{SWB}  \\ \cline{4-7}
    & & & dev & test & Hub5'00 & Hub5'01 \\ \hline
    \multicolumn{3}{|c|}{baseline w/o DMAO} & 7.7 & 7.9 & 14.0 & 12.8  \\ \hline
    \multirow{6}{*}{20\%} & 0.1 & \multirow{2}{*}{1} & 7.6 & \textbf{7.7}  & 13.4 & 12.3 \\ \cline{2-2} \cline{4-7}
    & \multirow{3}{*}{0.15} & & \textbf{7.3} & 7.9 & 13.5 & 12.3 \\ \cline{3-7}
    & & 4 & 7.5 & 8.0 & 13.5 & 12.3  \\ \cline{3-7}
    & & 8 & 7.5 & 7.9 & 13.9 & 12.3   \\ \cline{2-7}
    & 0.2 & \multirow{5}{*}{1} & 7.4 & \textbf{7.7} & 13.5 & 12.5 \\ \cline{2-2} \cline{4-7}
    & 0.25 & & 7.5 & \textbf{7.7} & \textbf{13.3} & \textbf{12.2} \\ \cline{1-2} \cline{4-7}
    10\% & \multirow{3}{*}{0.15} & & 7.7 & 8.0 & 13.5 & 12.5 \\ \cline{1-1} \cline{4-7}
    30\% & & & 7.7 & 8.1 & 14.0 & 12.5 \\ \cline{1-1} \cline{4-7}
    50\% & & & 8.0 & 7.9 & 14.0 & 13.0 \\ \hline
  \end{tabular}}
  \vspace{-1em}
\end{table}

\subsubsection{Applying DMAO to E-branchformer}
To validate that DMAO can enhance model capacity regardless of the architecture,
we also apply DMAO to the E-branchformer model and present the results in
Table~\ref{tab:DMAO_E_branformer}.
The results align with those obtained using the Conformer, showing that DMAO
generally enhances the performance of the E-branchformer encoder across various
model sizes and datasets.

\begin{table}
  \centering
  \setlength{\tabcolsep}{0.37em}
  \caption{WERs [\%] of applying DMAO on E-branchformer across
   different model sizes. The same DMAO schedule from Table~\ref{tab:overall_results_conformer}
   is used, with DMAO applied once at 20\% (15\% for \textit{LBS}) of total training steps,
   an adaptation ratio of $\delta$=0.15, and the first-order Taylor approximation as the metric.}
   \label{tab:DMAO_E_branformer}
  \scalebox{0.9}{\begin{tabular}{|c|c|c|c|c|c|}
    \hline
    \multirow{2}{*}{\shortstack{Params.\\ $[\text{M}]$}} & \multirow{2}{*}{DMAO}
     & \multicolumn{2}{c|}{SWB} & \multicolumn{2}{c|}{LBS} \\ \cline{3-6}
    & & Hub5'00 & Hub5'01 & dev-other & test-other\\ \hline\hline
    \multirow{2}{*}{25.7} & no & 13.9 & 12.8 & 7.3 & 7.7 \\ \cline{2-6}
    & yes & \textbf{13.8} & \textbf{12.5} & \textbf{6.9} & \textbf{7.4} \\ \hline \hline
    \multirow{2}{*}{56.9} & no & 13.6 & 12.1 & 6.6 & 6.9 \\ \cline{2-6}
    & yes & \textbf{13.0} & \textbf{11.7} & \textbf{6.5} & \textbf{6.8} \\ \hline \hline
   \multirow{2}{*}{100.2} & no & \textbf{12.9} & \textbf{11.9} & 6.3 & 6.7 \\ \cline{2-6}
    & yes & 13.0 & \textbf{11.9} & \textbf{6.2} & \textbf{6.6} \\ \hline
  \end{tabular}}
\end{table}

\subsubsection{Updated Architecture Analysis}
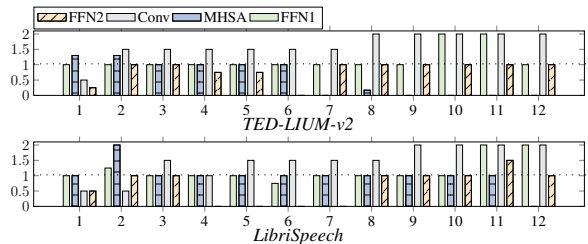
\begin{figure}
  \captionsetup[subfloat]{position=bottom,labelformat=empty, belowskip=-1pt,
  aboveskip=-1pt, font=scriptsize}
  \centering
  \subfloat[TED-LIUM-v2]{
  \input{tikzpicture/model_1.tex}
  }
  \vspace{0mm}
  \subfloat[LibriSpeech]{
  \input{tikzpicture/model_3.tex}
  }
  \vspace{-1em}
  \caption{Distribution of model parameters with DMAO$^\ddag$ (shown in Table~\ref{tab:overall_results_conformer}) across all depths. The y-axis represents the ratio of parameters after vs. before DMAO for each module. The dotted line indicates the baseline$^\dag$ (shown in Table~\ref{tab:overall_results_conformer}).}
  \label{fig:ratio_after_update}
  \vspace{-4mm}
\end{figure}

\begin{figure}
  \centering
  \includegraphics[scale=0.17]{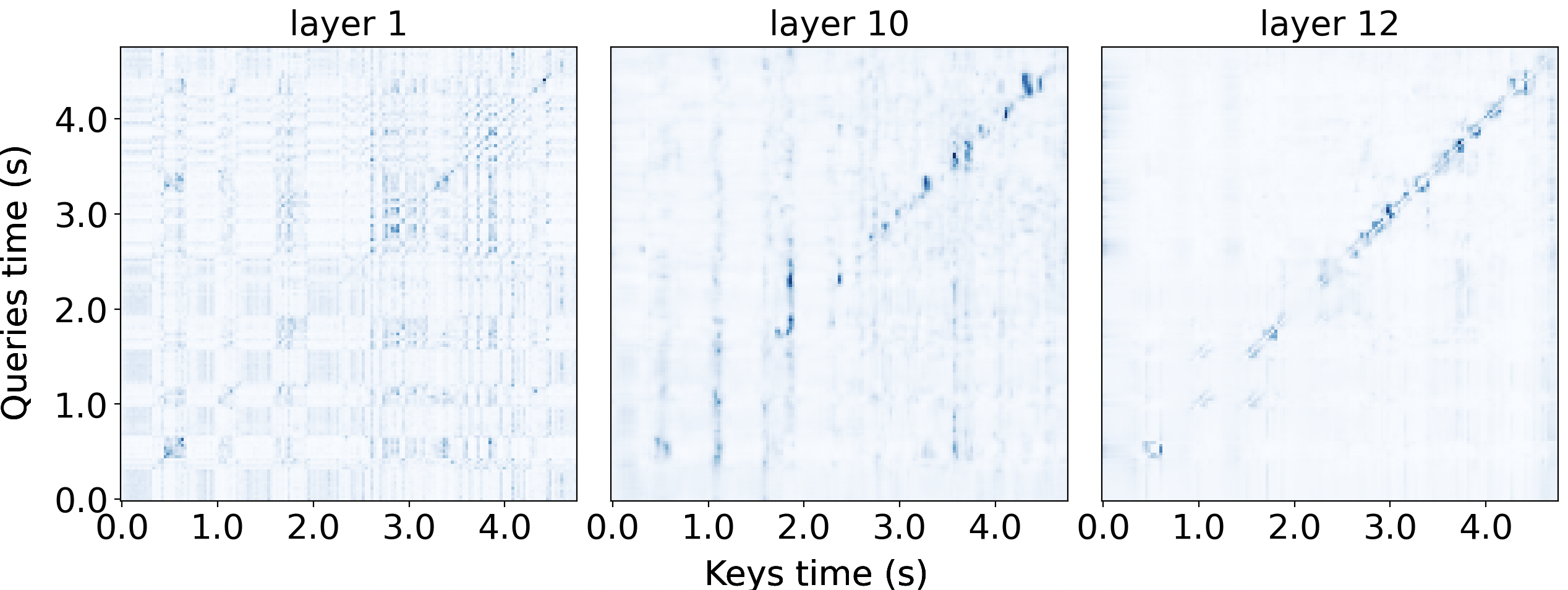}
  \vspace{-1em}
  \caption{Averaged self-attention scores across the 6 attention heads in
  the 1st, 10th, and 12th layers of a randomly selected sequence from \textit{LBS}
   for the baseline$^\dag$ (from Table~\ref{tab:overall_results_conformer}).}
  \label{fig:attn_score_plot}
  \vspace{-1em}
\end{figure}

Figure \ref{fig:ratio_after_update} illustrates the distribution of parameters
in the model after DMAO.
We observe that in the lower layers, more MHSA is utilized, while fewer Convs
are employed.
In contrast, in the top layers, MHSA heads are removed, and more Convs
 and FFNs are introduced.
To explain this, we illustrate Figure \ref{fig:attn_score_plot} to show that as
 the depth increases, the attention maps display highly diagonal patterns.
Similar observations have also been made in works
\cite{zhang2021diagonal2, zhang2021diagonal,burchi2021conformer, kim2022squeezeformer} for the ASR task.
As the range of learned context grows with increasing depth, the global view of
the entire sequence appears to become less useful for the upper layers.
Moreover, \cite{shim2022role} reveals that attention roles can be divided into
phonetic and linguistic localization.
The lower layers primarily perform phonetic localization, where the model focuses
on content-wise similar frames, which may be farther apart, to extract
phonologically meaningful features.
In contrast, higher layers focus on neighboring frames and aggregate the
information for text transcription.
This helps explain our observation, as Convs and FFNs are effective for local
processing, while MHSA is specialised for global interaction.
Additionally, we observe that more MHSA heads are used for \textit{LBS} than
 for \textit{TED-v2}.
A possible explanation is that \textit{LBS} has longer sequences than
\textit{TED-v2} (average 12.3s vs 8.2s), so more MHSA heads are needed to
capture dependencies over greater distances.

\section{Conclusion}
We propose the DMAO training framework to dynamically optimize
the acoustic encoder architecture during training.
The optimization is achieved through the grow-and-drop paradigm, where the
model's parameter allocation is balanced by redistributing parameters from
underutilized regions to areas where they are most needed, thereby enhancing the
model's capacity.
We demonstrate the effectiveness of DMAO by applying it with a CTC encoder
on various datasets, model sizes, and architectures (Conformer, E-Branchformer).
The results show a consistent improvement of up to $\sim6\%$ relative to the baseline when applying DMAO, with only negligible training overhead.

\section{Acknowledgement}
\scriptsize
This work was partially supported by NeuroSys, which as part of the
initiative “Clusters4Future” is funded by the Federal Ministry of
Education and Research BMBF (funding ID 03ZU2106DD),
and by the project RESCALE within the program \textit{AI Lighthouse
Projects for the Environment, Climate, Nature and Resources} funded by
the Federal Ministry for the Environment, Nature Conservation, Nuclear
Safety and Consumer Protection (BMUV), funding IDs: 6KI32006A and 6KI32006B.

\bibliographystyle{IEEEtran}
\bibliography{mybib}

\end{document}

%% file: tikzpicture/model_1.tex
\pgfplotstableread{
Label FFN1 Conv MHSA FFN2
1 1 0.5 1.3 0.25
2 1 1.5 1.3 1
3 1 1.5 1 1
4 1 1.5 1 0.75
5 1 1.5 1 0.75
6 1 1.5 1 0
7 1 1.5 0 1
8 1 2 0.17 1
9 1 2 0 1
10 2 2 0 1
11 2 2 0 1
12 1 2 0 1
}\data

\begin{tikzpicture}[scale=0.47]
  \tikzstyle{every node}=[font=\large]
    \begin{axis}[
        ybar,
        bar width=5,
        width=\textwidth,
        height=.2\textwidth,
        ymin=0,
        ymax=2.1,
        xtick=data,
        legend columns=4,
        legend style={at={(0,1.2)},anchor=west, row sep=0.5cm},
        reverse legend=true,
        xticklabels from table={\data}{Label},
        xticklabel style={text width=3cm,align=center},
        my ybar legend,
    ]
        \addplot [fill=YellowGreen!30]
            table [y=FFN1, meta=Label, x expr=\coordindex]
                {\data};
                    \addlegendentry{FFN1}
        \addplot [fill=NavyBlue!30!, postaction={pattern={Lines[angle=0,
        distance=1.5mm, line width=0.03mm]}}]
            table [y=MHSA, meta=Label, x expr=\coordindex]
                {\data};
                    \addlegendentry{MHSA}

        \addplot [fill=Gray!20]
            table [y=Conv, meta=Label, x expr=\coordindex]
                {\data};
                    \addlegendentry{Conv}
        \addplot [fill=Dandelion!30,postaction={pattern={Lines[angle=45,
        distance=1mm, line width=0.03mm]}},]
            table [y=FFN2, meta=Label, x expr=\coordindex]
                {\data};
                \addlegendentry{FFN2}
    \end{axis}

\draw[thin, dotted] (0,0.89) -- (15.5,0.89); 

\end{tikzpicture}

%% file: tikzpicture/model_3.tex
\pgfplotstableread{
Label FFN1 Conv MHSA FFN2
1 1 0.5 1 0.5
2 1.25 0.5 2 1
3 1 1.5 1 1
4 1 1 1 0
5 1 1.5 1 0
6 0.75 1.5 1 0
7 1 1.5 1 0
8 1 1.5 1 1
9 1 2 1 1
10 1 2 1 1
11 2 2 1 1.5
12 2 2 0 1
}\data

\begin{tikzpicture}[scale=0.47]
  \tikzstyle{every node}=[font=\large]
    \begin{axis}[
        ybar,
        bar width=5,
        width=\textwidth,
        height=.2\textwidth,
        ymin=0,
        ymax=2.1,
        xtick=data,
        legend columns=4,
        legend style={at={(0,1.2)},anchor=west, row sep=0.5cm},
        reverse legend=true,
        xticklabels from table={\data}{Label},
        xticklabel style={text width=3cm,align=center},
        my ybar legend,
    ]
        \addplot [fill=YellowGreen!30]
            table [y=FFN1, meta=Label, x expr=\coordindex]
                {\data};
        \addplot [fill=NavyBlue!30!, postaction={pattern={Lines[angle=0,
        distance=1.5mm, line width=0.03mm]}}]
            table [y=MHSA, meta=Label, x expr=\coordindex]
                {\data};

        \addplot [fill=Gray!20]
            table [y=Conv, meta=Label, x expr=\coordindex]
                {\data};
        \addplot [fill=Dandelion!30,postaction={pattern={Lines[angle=45,
        distance=1mm, line width=0.03mm]}},]
            table [y=FFN2, meta=Label, x expr=\coordindex]
                {\data};
    \end{axis}

\draw[thin, dotted] (0,0.89) -- (15.5,0.89); 

\end{tikzpicture}